\documentclass[10pt, conference]{IEEEtran}

\usepackage{booktabs}
\usepackage{cite}
\usepackage{stix}
\usepackage{amsmath,amsfonts}
\usepackage{algorithmic}
\usepackage{graphicx}
\usepackage{subcaption}
\usepackage{textcomp}
\usepackage{xcolor}
\usepackage{float}
\usepackage{caption}
\usepackage{array}
\usepackage{fancybox}
\usepackage{listings}
\usepackage{color}
\usepackage{todonotes}
\usepackage{enumitem}
\usepackage{hyperref}
\usepackage{siunitx}
\usepackage[utf8x]{inputenc}

\graphicspath{{./img/}}

\definecolor{lightgray}{rgb}{.9,.9,.9}
\definecolor{darkgray}{rgb}{.4,.4,.4}
\definecolor{purple}{rgb}{0.65, 0.12, 0.82}
\definecolor{greenselected}{rgb}{.1643,.6715,.1643}
\definecolor{redselected}{rgb}{.92,.04,.04}

\newcommand{\green}[1]{\text{\colorbox{green!15}{\makebox[1em]{\strut$#1$}}}}
\newcommand{\greentab}[1]{\text{\colorbox{green!15}{\makebox[2em]{\strut$#1$}}}}
\newcommand{\red}[1]{\text{\colorbox{red!30}{\makebox[1em]{\strut$#1$}}}}
\newcommand{\greenselected}[1]{\text{\colorbox{greenselected}{\makebox[1em]{\strut$#1$}}}}
\newcommand{\redselected}[1]{\text{\colorbox{redselected}{\makebox[1em]{\strut$#1$}}}}
\newcommand{\blue}[1]{\text{\colorbox{blue!20}{\makebox[1em]{\strut$#1$}}}}
\newcommand{\gray}[1]{\text{\colorbox{gray!30}{\makebox[1em]{\strut$#1$}}}}
\newcommand{\grayreturn}[1]{\text{\colorbox{gray!30}{\makebox[4em]{\strut$#1$}}}}
\newcommand{\grayvalue}[1]{\text{\colorbox{gray!30}{\makebox[3.2em]{\strut$#1$}}}}

\newcommand{\projectname}[1]{\textsc{#1}}
\newcommand{\StyleAnalyzer}{\projectname{style-analyzer}}
\newcommand{\Lookout}{\projectname{Lookout}}
\newcommand{\Babelfish}{\projectname{Babelfish}}
\newcommand{\Codebuff}{\projectname{CodeBuff}}
\newcommand{\Naturalize}{\projectname{Naturalize}}
\newcommand{\Github}{GitHub}

\makeatletter
\newenvironment{CenteredBox}{%
\begin{Sbox}}{
\end{Sbox}\centerline{\parbox{\wd\@Sbox}{\TheSbox}}}
\makeatother

\lstdefinelanguage{JavaScript}{
  keywords={typeof, new, true, false, catch, function, return, null, catch, switch, var, if, in, while, do, else, case, break},
  keywordstyle=\color{blue}\bfseries,
  ndkeywords={class, export, boolean, throw, implements, import, this},
  ndkeywordstyle=\color{darkgray}\bfseries,
  identifierstyle=\color{black},
  sensitive=false,
  comment=[l]{//},
  morecomment=[s]{/*}{*/},
  commentstyle=\color{purple}\ttfamily,
  stringstyle=\color{red}\ttfamily,
  morestring=[b]',
  morestring=[b]"
}

\begin{document}


\def \whitespace {\mathvisiblespace}
\def \tab {\mapsto}
\def \newline {\rotatebox[origin=c]{180}{$\Rsh$}}
\def \spaceinc{\mathvisiblespace^+}
\def \spacedec{\mathvisiblespace^-}
\def \tabinc {\mapsto^+}
\def \tabdec {\mapsto^-}
\def \singlequote {\textquotesingle}
\def \doublequote {\textquotedbl}
\def \noop {\varnothing}
\def \longtab {\longmapsto}
\def \longtabinc {\longmapsto^+}
\def \longtabdec {\longmapsto^-}

\def \precisionsTopRepos {95}
\def \numberTopRepos {19}
\def \minSamplesLeaf {80}
\def \uastCheckSpeedUp {1.5x }
\def \sizeThreshold {2 MB}
\def \parallelizeUastCheckSpeedUp {50\%}
\def \numberRulesAxios {183 }
\def \numberRulesJquery {480 }
\def \numberMistakesAxios{67 }
\def \numberMistakesJquery{106 }

\title{\StyleAnalyzer: fixing code style inconsistencies with interpretable unsupervised algorithms}

\author{\IEEEauthorblockN{Vadim Markovtsev, Waren Long, Hugo Mougard, Konstantin Slavnov, Egor Bulychev}
\IEEEauthorblockA{\textit{source\{d\}} \\
Madrid, Spain \\
\{vadim,waren,hugo,konstantin,egor\}@sourced.tech}
}

\maketitle

\begin{abstract}
Source code reviews are manual, time-consuming, and expensive.
Human involvement should be focused on analyzing the most relevant aspects of the program,
such as logic and maintainability, rather than amending style, syntax, or formatting defects.
Some tools with linting capabilities can format code automatically and report various stylistic
violations for supported programming languages.
They are based on rules written by domain experts, hence, their configuration is often tedious,
and it is impractical for the given set of rules to cover all possible corner cases.
Some machine learning-based solutions exist, but they remain uninterpretable black boxes.

This paper introduces \StyleAnalyzer, a new open source tool to automatically fix code formatting violations
using the decision tree forest model which adapts to each codebase and is fully unsupervised.
\StyleAnalyzer{} is built on top of our novel assisted code review framework, \Lookout. 
It accurately mines the formatting style of each analyzed Git repository and expresses the found format patterns
with compact human-readable rules.
\StyleAnalyzer{} can then suggest style inconsistency fixes in the form of code review comments.
We evaluate the output quality and practical relevance of \StyleAnalyzer{} by demonstrating that it can reproduce the original style with high precision, measured on \numberTopRepos{} popular JavaScript projects, and by showing that it yields promising results in fixing real style mistakes.
\StyleAnalyzer{} includes a web application to visualize how the rules are triggered.
We release \StyleAnalyzer{} as a reusable and extendable open source software package on \Github{} for the benefit of the community.
\end{abstract}


\section{Introduction}
\label{sec:introduction}

The way source code is formatted has a significant impact on both comprehensibility and maintainability~\cite{miara, hindle}, and ensuring that a codebase is consistent in style is tedious and time consuming.
When human reviewers have to consider formatting during a code review session, their ability to spot defects and bugs is diluted by too many concerns~\cite{ineffective_code_reviews}.
At Google, software engineers who obtain \textit{readability} certification for a specific language are assigned for approval during the review process to ensure stylistic consistency across the codebases~\cite{google}.
However, those developers in charge of a particular codebase might rotate on a regular basis, and some projects might not even have a linting process configured early on.
Furthermore, different programmers often have different code formatting preferences, such as indentation, white space usage, or brace positioning. This eventually leads to variations of the resulting code style.
Of course, there are plenty of configurable linting tools for source code, whether included into IDEs like \projectname{IntelliJ} or external applications like \projectname{ESLint} for JavaScript.
However, configuring the style checks is not obvious; these tools can be too opinionated and hard to set up to satisfy team's wishes.
Furthermore, there may be dependencies between options, and the tools struggle to take the context information into account~\cite{veerman}.
Finally, linters do not always suggest fixes for the violated rules, and when they do, applying those fixes might not be convenient for programmers.

In this paper, we introduce \StyleAnalyzer{}, a tool to solve the automatic code formatting problem at code review time, suggesting changes to pull requests on \Github{} when style inconsistencies are detected.
An example of such a suggestion is given in Figure~\ref{fig:github-suggested-change}.
We include this solution into our new assisted code review framework, \Lookout, that allows running pluggable code analyzers over pull requests.
Thus our training set is restricted to a single repository.
In order to make the suggested changes interpretable and establish trust with the users, we employ an adaptive machine learning model which learns code formatting rules from existing code.
We gather the implicit and explicit user feedback on the triggered rules and have the ability to disable the misbehaving ones while leaving the rest intact.
Finally, the analysis performance is high enough to yield results with small time footprint - under five minutes for large pull requests.
To satisfy our requirements and limitations, we first define the underlying code style of a repository in terms of language models, then train a decision tree forest on the AST-augmented token stream.
Finally, we extract rules from the trees and optimize them.
Debugging such a complex pipeline requires dedicated tooling, so we developed a benchmarking suite and an application to visualize how the rules are triggered.

The main contributions of this paper are:
\begin{itemize}[label={$\bullet$}]
\item \StyleAnalyzer{}, an analysis running on \Lookout{} that mines interpretable code formatting rules using machine learning, validates new code against them, and suggests fixes when appropriate.
\item A new assisted code review framework, \Lookout, which watches \Github{} repositories and triggers a set of analyses when pull requests are updated or new commits are pushed. \Lookout{} reports the result of these analyses as \Github{} comments to pull requests, leveraging the recently appeared \emph{\Github{} Suggested Changes} feature. \Lookout{} allows for rapid development of new analyses and provides a universal code parsing API.
\item A web application which annotates the analyzed source code with the extracted features of code tokens and the triggered rules.
\end{itemize}

The rest of the paper is organized as follows.
Section~\ref{sec:related-work} revises prior work related to fixing formatting defects.
In section~\ref{sec:methodology}, we explain the model behind \StyleAnalyzer.
Section~\ref{sec:implementation} presents \Lookout, our framework for assisted code review and details how it helped to implement \StyleAnalyzer.
In section~\ref{sec:evaluation}, we detail the evaluation of our model.
Section~\ref{sec:future-work} first discusses the current limitations of \StyleAnalyzer{} approach and then explains how they can be addressed in future works.

\section{Related Work}
\label{sec:related-work}

Some machine learning approaches have already been explored to suggest code improvements for stylistic consistency.
Allamanis \textit{et al.} were the first to introduce the coding convention inference problem.
They built \Naturalize, a language-agnostic code formatting suggester that learns formatting conventions and generates rules from them~\cite{naturalize}.
\Naturalize{} proceeds in two steps: first it generates formatting candidates, second it ranks those candidates for later use in development tooling if they meet a quality threshold.
Nevertheless, \Naturalize{} considers only the local context and can produce semantically disruptive suggestions; thus it is not fully automatic.
More recently, \Codebuff{} ---~an automatic code formatter for Java, SQL and ANTLR~--- was proposed~\cite{codebuff}.
It exerts machine learning to obtain abstract formatting rules from a representative corpus.
However, \Codebuff{} needs to be configured by example and covers neither mixed indentation (tabs vs. spaces) nor mixed quotes (single vs. double).
Besides, \Codebuff{} uses a \textit{k}-Nearest Neighbor machine learning model to classify the unknown feature vectors, which makes its decisions harder to understand by humans.


Another closely related group of tools encompasses rule-based \textit{default formatters}. In contrast to \StyleAnalyzer{} that learns the existing style of a codebase, they are often opinionated and enforce a specific subset of styles approved by their designers: if a user is targeting a specific formatting style, those tools might prove impossible to setup. Among the most popular ones, we can point out \projectname{GNU~indent}\footnote{\url{https://www.gnu.org/software/indent/}} and \projectname{Prettier}\footnote{\url{https://github.com/prettier/prettier}} for example.

Other approaches address slightly different albeit related issues.
As an illustration, Wang \textit{et al.} have developed a heuristic solution to automatically insert blank lines in methods to improve their readability~\cite{wang}.
\section{Methodology}
\label{sec:methodology}

\newcounter{tablecounter}
\newcounter{figurecounter}

\begin{figure}[t]
\setcounter{figure}{\value{figurecounter}}
\refstepcounter{figurecounter}
\centering
\includegraphics[width=1\linewidth]{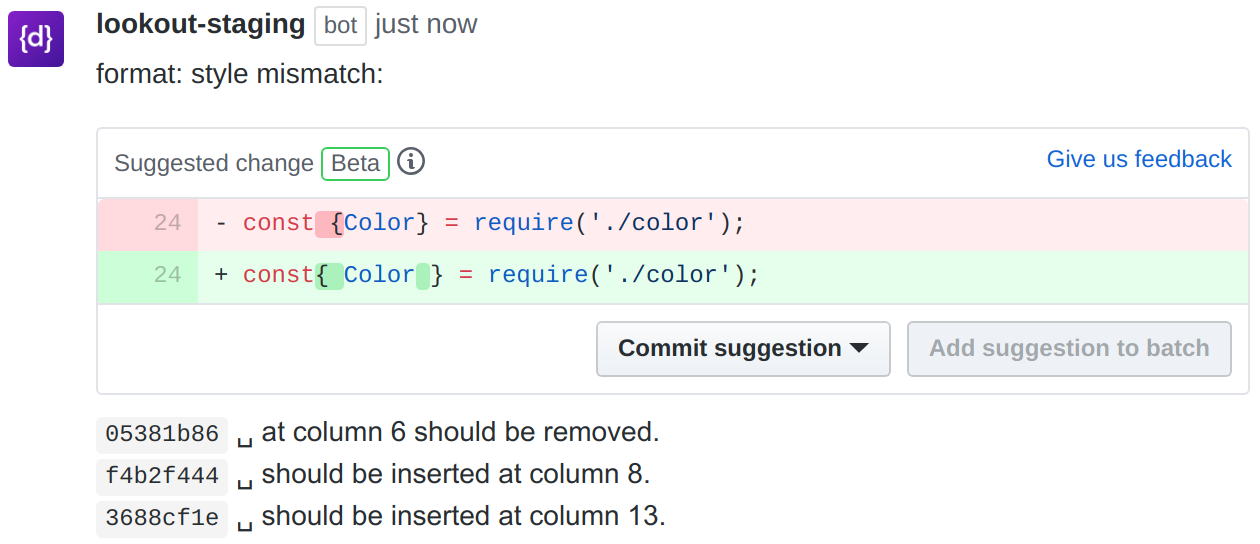}
\vspace*{-0.5em}
\captionof{figure}{Example of \Lookout{} comment output from \StyleAnalyzer{} and published as a \emph{Github Suggested Change}.}
\vspace*{-1em}
\label{fig:github-suggested-change}
\end{figure}

\StyleAnalyzer{} learns the underlying code formatting rules of a repository in a completely unsupervised manner ---~and with zero prior domain knowledge~--- from the supplied files constituting a specific Git revision.
\StyleAnalyzer{} is built on the concept of a holistic language model of source code. Language models have been largely explored to calculate various probability distributions of source code using maximum likelihood estimates~\cite{nguyen}. We do not rely on any explicitly labeled data; instead, we model the formatting code tokens from the surrounding context where they occur in the files. We further model the absence of the formatting tokens in the rest of the relevant places (in-between semantic tokens). The learned language model predicts whether new, proposed code changes follow the established conventions.

\subsection{Code style and feature extraction}
\label{subsec:methodology:code-style-feature-extraction}

We represent a source code file as a linear sequence of tokens. An example of representation is given in Figure~\ref{fig:snippet}.
Some of these tokens correspond to the nodes in the AST representing the file, while others are inserted to mirror whitespace characters, keywords, quotes, and braces ---~in other words, auxiliary tokens which are absent from the AST but allow the accurate reconstruction of the original file.
The exact set of the latter depends on the programming language and the AST structure; we stick to JavaScript and use \Babelfish{}~\cite{bblfsh} to obtain ASTs with a structure that is common to many languages. We will refer to those ASTs as UASTs for \emph{Universal Abstract Syntax Trees}.
The ability to reproduce the exact source code from the parsed stream of tokens is a prerequisite.

Regarding the formatting elements which we predict, we consider 9 atomic classes listed in Table~\ref{fig:classes}.
To fit our sequence prediction framework, those classes are combined to form compound sequences, e.g., four spaces indentation increase ($\mathvisiblespace^+ \mathvisiblespace^+ \mathvisiblespace^+ \mathvisiblespace^+$) or two consecutive newlines (\rotatebox[origin=c]{180}{$\Rsh$} \rotatebox[origin=c]{180}{$\Rsh$}).

We take advantage of the sequence prediction framework to correctly model insertions and deletions.
From a given sample, the prediction to a larger sequence can model insertion, to a smaller sequence, deletion. To allow insertions from places where there is no formatting and full deletions from places where formatting is present, we include the empty sequence as a label.

\newcolumntype{M}[1]{>{\raggedright}m{#1}}
\begin{figure}[t]
\begin{tabular}{l M{0.2cm} l}
\toprule
1. & $\whitespace$ & ~ whitespace \tabularnewline
2. & $\tab$ & ~ tabulation  \tabularnewline
3. & $\newline$ & ~ newline  \tabularnewline
4. & $\spaceinc$ & ~ whitespace indentation increase \tabularnewline
5. & $\spacedec$ & ~ whitespace indentation decrease  \tabularnewline
6. & $\tabinc$ & ~ tabulation indentation increase \tabularnewline
7. & $\tabdec$ & ~ tabulation indentation decrease  \tabularnewline
8. & \singlequote & ~ single quote  \tabularnewline
9. & \doublequote & ~ double quote  \tabularnewline
10. & $\noop$ & ~ empty gaps between non-label nodes, NOOP \tabularnewline
\bottomrule
\end{tabular}
\setcounter{figure}{\value{tablecounter}}
\captionof{table}{List of atomic label classes}
\vspace*{-1em}
\label{fig:classes}
\end{figure}

\begin{figure*}[t]
    \setcounter{figure}{\value{figurecounter}}
    \begin{CenteredBox}
    \begin{lstlisting}[frame=tb, framexleftmargin=0.05\textwidth, framexrightmargin=-.32\textwidth, mathescape=true, language=]
$\green{\noop}$function$\green{\whitespace}$classesToArray$\green{\noop}$($\green{\whitespace}$value$\green{\whitespace}$)$\green{\whitespace}${$\green{\newline}$
$\greentab{\longtabinc}~~$if$\green{\whitespace}$($\green{\whitespace}$isArray$\green{\noop}$($\green{\whitespace}$value$\green{\whitespace}$$\gray{)}$$\greenselected{\whitespace}$$\gray{)}$$\greenselected{\whitespace}$$\gray{\{}$$\redselected{\noop}$$\grayreturn{\mbox{return}}$$\greenselected{\whitespace}$$\grayvalue{\mbox{value}}$$\greenselected{\noop}$$\gray{;}$$\red{\noop}$}$\green{\newline}$
$\hspace{3.2em}$if$\green{\whitespace}$($\green{\whitespace}$typeof$\green{\whitespace}$value$\green{\whitespace}$===$\green{\whitespace \mbox{"}}$string$\green{\mbox{"}\whitespace}$)$\green{\whitespace}${$\green{\newline}$
$\greentab{\longtabinc}~~~~~~$return$\green{\whitespace}$value$\green{\noop}$.$\green{\noop}$ match($\green{\whitespace}$rnothtml$\green{\whitespace}$)$\green{\whitespace}$||$\green{\whitespace}$[]$\green{\noop}$;$\green{\newline}$
$\greentab{\longtabdec}~~$}$\green{\newline}$
$\hspace{3.2em}$return$\green{\whitespace}$[]$\green{\noop}$;$\green{\newline}$
$\greentab{\longtabdec}$}$\green{\noop}$
    \end{lstlisting}
    \end{CenteredBox}
    \vspace*{0.5em}
    \captionof{figure}{Example of JavaScript code with style inconsistencies. The source code is annotated according to our code representation. The green labels fit the model's predictions while the red ones differ, and the highlighted labels correspond to the 10-token window used for feature extraction around the first red sample.}
    \refstepcounter{figurecounter}
    \label{fig:snippet}
    \vspace*{1.5em}
    \begin{subfigure}[b]{1\textwidth}
        \begin{minipage}{.5\textwidth}
            \begin{itemize}
                \setlength\itemsep{0.5em}
                \item[] \textbf{y :} $\noop$
                \item[] \textbf{\^{y} :} $\newline \tabinc$
                \item[] Confidence: 0.975
                \item[] Support: 3230
                \item[] \textbf{Attributes}
                    \begin{itemize}
                        \setlength\itemsep{0.1em}
                        \item[]\texttt{ -1.value} = \texttt{\{}
                        \item[]\texttt{ -3.value} not in \{\texttt{else}\}
                        \item[]\texttt{ +1.internal\_type} not in \{\texttt{StringLiteral}\}
                        \item[]\texttt{ +2.roles} not in \{\texttt{LITERAL}, \texttt{COMMENT}\}
                        \item[]\texttt{ \^{ }1.roles} in \{\texttt{IF}, \texttt{STATEMENT}\}
                    \end{itemize}
            \end{itemize}
            \captionof{figure}{}
            \label{fig:rule}
        \end{minipage}
        \vline
        \begin{minipage}{.5\textwidth}
            \begin{itemize}
                \setlength\itemsep{0.5em}
                \item[] \textbf{Features}
                \begin{itemize}
                    \item[] \textbf{root:} \{\} 4 items
                    \vspace*{0.1em}
                    \begin{itemize}
                        \setlength\itemsep{0.1em} 
                        \item[] \textbf{left:} [] 5 items
                        \item[] \textbf{node:} [] 1 item
                        \item[] \textbf{parents:} [] 2 items
                        \begin{itemize}
                            \item[] \textbf{1:} \{\} 2 items
                            \item[] ~~~\textbf{internal\_type:} IfStatement
                            \item[] ~~~\textbf{roles:} \{IF, STATEMENT\}
                            \item[] \textbf{2:} \{\} 2 items
                        \end{itemize}
                        \item[] \textbf{right:} [] 5 items
                    \end{itemize}
                \end{itemize}
            \end{itemize}
            \captionof{figure}{}
            \centering
            \captionsetup{justification=centering}
            \label{fig:features}
        \end{minipage}
    \end{subfigure}
    \vspace*{0.2em}
    \caption[figure]{(a) Prediction of our model and insight into the triggered rule. The window is centered at the highlighted red label in Figure~\ref{fig:snippet}. The rule's details include the its confidence and support. (b) Features extracted at the same location in the code. For example, the first parent in the UAST hierarchy has the following two roles: \texttt{If} and \texttt{Statement}.}
    \label{fig:rule-and-features}
    \vspace*{-1em}
    \refstepcounter{figurecounter}
\end{figure*}

The filtered labels are one-hot encoded, and we solve a classification problem.
To extract features from the formatting elements, we mix sequential and structural information by combining 2 different representations of source code \cite{dipenta}. 
We explore a symmetric window of fixed size in the stream of tokens, as well as climb a few levels up the AST hierarchy.
As described in Figures \ref{fig:snippet} and \ref{fig:rule-and-features}, the window that we adopted in our experiments was 5 nodes to the left, 5 nodes to the right and 2 AST parents up.
A token's parent here means the \textit{Lowest Common Ancestor} (LCA) of the closest token's sibling in the UAST ---~in other words, the deepest UAST node that has both the closest left and right UAST nodes as descendants. To compute the LCA, we trace the UAST two times respectively from the two siblings down the root, and return the common node just before the encountered mismatch.

Universal Abstract Syntax Trees provide the following node attributes:

\begin{description}
    \item[Value] the content of the corresponding token, if it exists, otherwise an empty string.
    \item[Internal type] the string which indicates the type of the node in the native AST ---~that is, before the UAST conversion.
    \item[Roles] the set of strings which indicate UAST node roles. There is a list of supported roles in the official \Babelfish{} documentation with around 200 elements, e.g. \texttt{Identifier}, \texttt{Literal}, \texttt{Comment}.
    \item[Start position] the line number and the column number where the node begins. It matches the token start position if the node is an AST leaf, otherwise, it is defined recursively as the minimum start position among its children.
    \item[End position] the line number and the column number where the node ends. It is defined similarly to the start position.
\end{description}

Hence, we decided to record the following features for each formatting element:

\begin{description}
    \item[Label] matches the one-hot encoded label if the token is a formatting sequence. It is collected only for the nodes to the left of a label.
    \item[Internal type] one-hot encoded. If there is no corresponding UAST node, all zeros.
    \item[Reserved index] one-hot encoded ---~for those tokens with an empty internal type, we collect all possible token values in the analyzed files.
    \item[Length] the length of the token value.
    \item[Roles] a fixed length vector with "1"-s at the indexes of the relevant roles. Again, if the token is not backed up with a UAST node, it is all zeros.
    \item[Position information] collected only for the nodes to the left of a label. It is all zeros for the nodes to the right.
    \begin{itemize}
        \item File offset difference with the previous node in the sequence.
        \item Column number difference alike.
        \item Line number difference alike.
    \end{itemize}
\end{description}

Regarding the parents, we extract their internal types and roles. Besides, we record the start position of the predicted node, namely the line and column numbers.

One of the challenges of designing an unsupervised predictive model is to not leak the information about the labels to the engineered features.
It is easy to miss some obvious relations between the context and the label which are perilous for an unbiased reasoning at test time~\cite{leakage}.
For example, we could include the difference in offsets between the immediate left and right neighbors as one of the features.
This difference indicates the exact length of the predicted token in-between and the model inevitably overfits to it.
As a result, it becomes limited to suggestions of the same length, e.g. it can no longer predict zero-length NOOPs instead of non-empty formatting tokens.
We mitigated the aforementioned perils by selecting different features for the left and the right token siblings.
More precisely, the right token features are limited to semantic-only information to avoid any formatting information leakage.

The overall size of our feature vector sums to over 4000. We apply univariate feature selection~\cite{tang2014feature} to keep the 500 best features according to the ANOVA F-value criterion.

The number of features to compute grows rapidly with the size of the analyzed repository to reach tens of millions of floating point values for a large repository, and the feature extraction step becomes the main pipeline bottleneck. It is therefore important to tune its performance.  We took advantage of the efficient sparse data structure operations in the \projectname{scipy}~\cite{scipy} library to avoid unnecessary computations.
We saturate the size of the training set to guarantee the strict run time bounds. It is limited to the first \sizeThreshold{} of concatenated randomly shuffled files. We additionally filter out files that are automatically generated, e.g. minified, by the maximum allowed line length of 500.

\subsection{Decision Trees}
\label{subsec:methodology:decision-trees}

Given the extracted features, we opt for decision trees as our machine learning model. This choice is motivated by our interpretability ---~“white box”~--- requirement.
Decision trees naturally and transparently explain their predictions by following the respective branches~\cite{breiman}.
Decision trees group examples seen in the training set into different leaves which are created to minimize the diversity of examples in each leaf.
A single decision tree is unlikely to achieve good predictive power on plenty of data; hence, the common practice is to train several decision trees and combine their predictions.
There are two widespread approaches: ensembling (decision tree forest) and boosting (gradient boosted decision trees, GBDT).
The random tree forest algorithm~\cite{random_forests} better suits our requirements since GBDT are hard to interpret due to tree chaining.
We use the \texttt{RandomForestClassifier} implementation from \projectname{scikit-learn}~\cite{scikit-learn}.

We run hyper-parameters optimization during the training phase to maximize the accuracy metric. More precisely, we perform 100 iterations of Bayesian optimization using Gaussian Processes~\cite{skopt} with stratified 3-fold validation to optimize the following hyper-parameters:

\begin{description}
    \item[Model] Whether to train a random forest or a decision tree;
    \item[Depth] Authorized maximum depth for the trees;
    \item[Considered features] Features considered for each split during tree construction;
    \item[Minimum samples per split] Minimum number of samples to split a node during tree construction;
    \item[Minimum samples per leaf] Minimum number of samples in a leaf during tree construction;
\end{description}

100 iterations were always enough to converge for all the repositories in our evaluation dataset from subsection \ref{subsec:evaluation:style-modeling-benchmark}.

After obtaining a trained decision tree forest, we transform it to production rules~\cite{quinlan}.
According to the established terminology, we call feature comparisons along a branch attribute comparisons.
Each path from the root of a tree to one of its leaves is a set of attribute comparisons which leads to a specific label prediction.
We say that such a set of attribute comparisons is a rule.
Each rule has a training precision value, \emph{i.e.} the number of times it predicted correctly divided by the number of times its attribute comparisons were simultaneously true. We call this value the confidence of a rule.

Having extracted the rules from the decision tree forest, we simplify them and reduce their number so that they are easier to comprehend.
First we set a confidence threshold and remove all the rules which are not precise enough.
If this threshold is close to 1, we end up with a small number of rules which are very precise but cover only a tiny fraction of the examples.
In contrast, if this threshold is close to 0, we keep all the rules, staying imprecise but retaining the best recall.

In order to simplify the rules, we merge all comparisons of the same attribute together, therefore doing at most two comparisons per attribute.

Afterwards, we perform attribute pruning ---~removing parts of the rules that are redundant and suppress generality. For each rule, we collect the sets of samples on which attribute comparisons \textit{incorrectly} predict the corresponding label. We further build the undirected similarity graph of those sets. It's vertices are attributes, and each pair is connected if the Jaccard similarity between their sample sets is bigger than a certain threshold (we chose 0.98). Finally, we perform Multilevel community detection \cite{multilevel} on that graph and greedily leave only a single representative of each community. The described pruning algorithm is computationally expensive, so we do not run it while optimizing the model hyper-parameters.

The last stage is removing duplicated rules after attribute pruning.
There also exist techniques to globally optimize a collection of rules~\cite{quinlan} but we found them impractically slow for our task.

The resulting set of production rules comprises the final model and is both considerably simpler than the raw decision trees, more accurate and works better with samples outside of the training set without sacrificing the recall.

%

\subsection{Applying the rules}
\label{subsec:methodology:applying-rules}

Application of the rules to changed code in pull requests has several points worth noting.
We create suggestions for every prediction that does not match the factual label.
It is possible for more than a single rule to fire ---~in such a case we use the most confident one.
It may also happen that no rule fires ---~then we silently proceed to the next sample.

There are paired format tokens, and they are processed independently according to our scheme.
Hence, there can be two predictions which contradict each other.
For instance, we may suggest a string literal with different left and right quotes.
We have to handle this special case separately by respecting the most confident rule and changing the paired prediction accordingly.

Besides, we check that the proposed fixes do not break the code.
For example, the model can remove a whitespace character in the middle of an expression and thus change the AST, as shown in Figure~\ref{fig:uast-check}.
We parse a new UAST for each changed line and compare it with the original UAST.
If any difference is detected, we drop the suggestions on the matching line.
However, this check is costly if there are many suggestions, and we leverage a parsing optimization: we avoid parsing a complete file each time.
Instead, we take the common parent UAST node of the checked line and parse the code block it spans over.
The parsing may fail ---~some nodes are not parsable in isolation; we then continue to climb up the AST hierarchy until we succeed.
On average, this optimization brings a \uastCheckSpeedUp speedup.
The AST checks are independent of each other, so they can be efficiently parallelized to leverage multiple CPU cores.
However, they execute in Python which has a Global Interpreter Lock, so the parallelization yields only a moderate \parallelizeUastCheckSpeedUp{} run time decrease on a 32-core machine ---~due to the large increase in computing power needed to obtain this reduction, we turn it off by default.
Similarly to training, we limit the size of changed files that we analyze in a pull request. Only the first \sizeThreshold{} are analyzed.

\StyleAnalyzer{} finishes analyzing a pull request by generating code for the suggestions. The main challenge here is to properly handle the indentation changes.
Given a sequence of altered lines, we may miss indentation fixes for some of them ---~that is, we might create indentation conflicts. There are two strategies:

\begin{enumerate}
    \item Independently generate code according to the predictions, ignoring the potential conflicts;
    \item Maintain a consistent state by also modifying the other impacted indentations for each prediction ---~\emph{e.g.}, increase the indentation of a whole block for a single prediction.
\end{enumerate}

The second strategy can be more accurate but would require a model that makes dependent predictions (such as a recurrent neural network or a hidden Markov model). \StyleAnalyzer{} makes independent predictions for each sample, as in the first approach.

\begin{figure}[t]
    \setcounter{figure}{\value{figurecounter}}
    \refstepcounter{figurecounter}
    \begin{subfigure}[b]{0.5\textwidth}
        \begin{lstlisting}[frame=tb, framexrightmargin=-0.03\textwidth, mathescape=true, language=]
    function classesToArray( value ) {
        if ( isArray( value ) ) {
            return value;
        }
        return [];
    }
        \end{lstlisting}
        \vspace*{-0.8em}
        \captionof{figure}{}
        \vspace*{0.3em}
    \end{subfigure}
    \begin{subfigure}[b]{0.5\textwidth}
        \begin{lstlisting}[frame=tb, framexrightmargin=-0.03\textwidth, mathescape=true, language=]
$\green{\noop}$function$\green{\whitespace}$classesToArray$\green{\noop}$($\green{\whitespace}$value$\green{\whitespace}$)$\green{\whitespace}${$\green{\newline}$
$\greentab{\longtabinc}~~$if$\green{\whitespace}$($\green{\whitespace}$isArray$\blue{\whitespace}$($\green{\whitespace}$value$\green{\whitespace}$)$\green{\whitespace}$)$\green{\whitespace}${$\green{\newline}$
$\greentab{\longtabinc}~~~~~~$return$\blue{\noop}$value$\green{\noop}$;$\green{\newline}$
$\greentab{\longtabdec}~~$}
$\hspace{3.2em}$return$\green{\whitespace}$[]$\green{\noop}$;$\green{\newline}$
$\greentab{\longtabdec}$}$\green{\noop}$
        \end{lstlisting}
        \vspace*{-0.8em}
        \captionof{figure}{}
        \vspace*{0.3em}
    \end{subfigure}
    \caption[figure]{(a) Original example of valid JavaScript code (b) examples of predictions of formatting elements that are changing the AST (in blue) and thus rejected during the AST filtering.}
    \vspace*{-1em}
    \label{fig:uast-check}
\end{figure}

\subsection{User feedback}
\label{subsec:methodology:user-feedback}

\begin{figure*}[t]
\setcounter{figure}{\value{figurecounter}}
\centering
\begin{minipage}{.5\textwidth}
\centering
\includegraphics[width=0.9\linewidth]{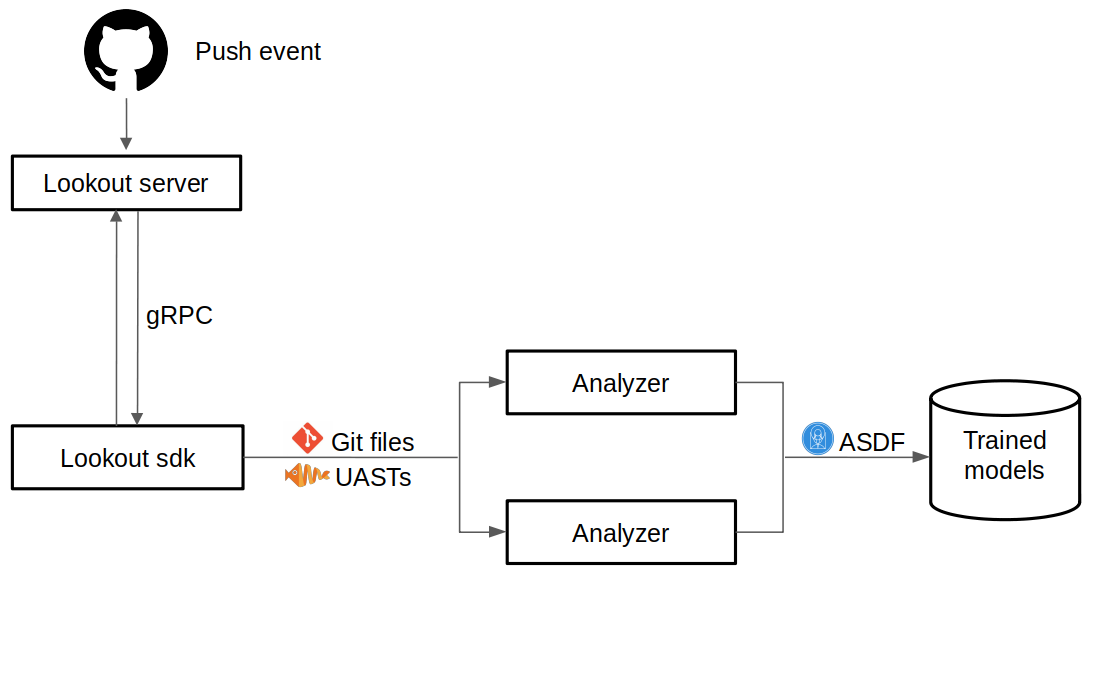}
\captionof{figure}{The flow of a \emph{Push} event}
\vspace*{-1em}
\refstepcounter{figurecounter}
\label{fig:push}
\end{minipage}%
\begin{minipage}{.5\textwidth}
\centering
\includegraphics[width=0.9\linewidth]{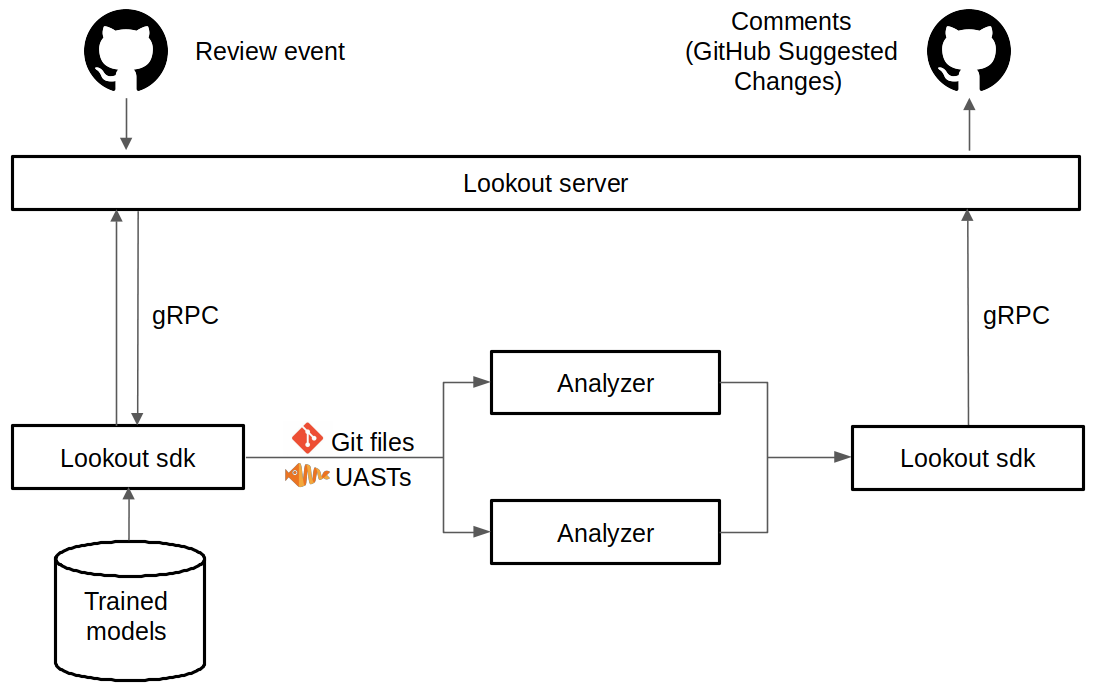}
\captionof{figure}{The flow of a \emph{Review} event}
\vspace*{-1em}
\refstepcounter{figurecounter}
\centering
\captionsetup{justification=centering}
\label{fig:review}
\end{minipage}
\end{figure*}

We hash each rule's body ---~the attribute comparisons and the predicted label~--- to a 32-bit unsigned integer.
Those hashes are prepended to fix descriptions in \Github{} comments, so users can identify which rules trigger.
Users have an ability to disable the rules which are wrong or too noisy in their opinion by blacklisting the identifiers in the \Lookout{} configuration file that resides at the repository root. Blacklisted suggestions are never shown.

Furthermore, \Lookout{} provides information about which suggestions have been merged and which have not.
We measure the ratio of unmerged suggestions over all suggestions for each rule.
Once the number of unmerged suggestions exceeds 10 and the ratio drops below 0.9, the rule gets blacklisted.

If we train another model, new rules may emerge. According to our calculations, a different random seed yields 75\% of unique rules and that makes our blacklisting mechanism less efficient. We mitigate this negative impact by only training new models each 100th commit. Better rule matching approaches can be explored in the future.

\section{Implementation}
\label{sec:implementation}

\subsection{\Lookout: an assisted code review framework}
\label{subsec:implementation:lookout}

\StyleAnalyzer{} runs its analysis when pull requests are updated on \Github{}.
There are other options to assist software developers: in an IDE while they type, in a Continuous Integration (CI) script or running analysis periodically as asynchronous scheduled jobs.
We discarded them for different reasons.
First, IDE solutions are expected to produce results instantly on developers' hardware and therefore enforce strict run time, memory and CPU utilization constraints.
Pull request analysis may run in the cloud on well-defined and powerful hosts and it is not required to generate results in few seconds, thus is potentially capable of more sophisticated assistance.
Second, CI does not have a good user interface for code suggestions.
In fact, the only way to report anything in CI is to print to standard output stream, whereas \Github{} enables single-click fix merges as can be seen in Figure~\ref{fig:github-suggested-change}.
Third, our tool should remain a part of the development workflow, otherwise there is a tendency to ignore or bypass the numerous accumulated suggestions.
This has been demonstrated at Google scale by Sadowski \textit{et al.}~\cite{lessons-static}.
Hence we abandon the scheduled jobs option.
Running the analysis during code review also has its drawbacks: the code review feedback loop is longer compared to IDEs because the main way to discover fixes is to create a pull request and look at \StyleAnalyzer's comments.

The purpose of the \Lookout{} framework is to deliver assisted code review to everybody in an easy-to-setup, easy-to-use, easy-to-extend fashion.
It contains the server application which listens to repository events from \Github{}. We call it the \Lookout{} server. It is possible to create applications which register with the \Lookout{} server to run code analyses. We call them \Lookout{} analyzers. Whenever new commits are pushed or pull requests are updated, the \Lookout{} server communicates with the registered analyzers. In case of a push event (see Figure~\ref{fig:push}), that communication is just a notification for analyzers to update their internal state, if they have one. In case of a pull request event (see Figure~\ref{fig:review}), the \Lookout{} server obtains a list of review suggestions from each of the analyzers. It aggregates the lists and posts comments to the corresponding lines in the files of the pull request.

If an analyzer suggests better code, it can format the Markdown text of the corresponding \Github{} comment in a special way to leverage \emph{\Github{} Suggested Changes}.
Released in October 2018, that feature provides a user interface to accept or reject proposed code edits with a single mouse click.
The \Lookout{} server inspects which code suggestions have been merged and supplies this information to analyzers, hence establishing a valuable user feedback loop.

\subsection{\Lookout{} architecture}
\label{subsec:implementation:architecture}

All the communication happens through gRPC remote procedure call interfaces.
The \Lookout{} server and each of the analyzers run as gRPC services.
This allows to scale easily and ensures solid isolation between the components.  
The \Lookout{} server is written in the Go programming language and is available online\footnote{\url{https://github.com/src-d/lookout}}.
Thanks to gRPC, there are no limitations on languages or frameworks for analyzer implementations.

The \Lookout{} server does not provide the contents of the changed files to analyzers by default.
Instead, it serves another gRPC endpoint for data retrieval.
It is possible for an analyzer to request the contents of any file at any revision in one of the watched Git repositories from that endpoint.
Besides, the \Lookout{} server is integrated with \Babelfish, the platform for language-agnostic source code parsing~\cite{bblfsh}.
\Babelfish{} parses code written in supported programming languages into Universal Abstract Syntax Trees (UASTs).
A UAST has the standard annotated format of the parsed syntax tree and simplifies cross-language code analysis by providing language-agnostic annotations (e.g. function, identifier) on the corresponding sub trees.

\Lookout{} abstracts analyzers from dealing with the \Github{} API, working with Git, parsing code, and discovering user feedback, thus allowing to focus on code analysis problems.

\subsection{Software Development Kit}
\label{subsec:implementation:software-development-kit}

\Lookout{} offers a Software Development Kit to ensure rapid development of new analyzers. There are two flavors of the SDK: low-level and high-level. The low-level SDK makes no assumptions on the nature of analyzers and serves as a thin proxy layer over gRPC. The high-level SDK
is based on the low-level one. It is written in Python, is also available online\footnote{\url{https://github.com/src-d/lookout-sdk-ml}} and targets stateful analyzers. Statefulness means here that a push event leads to a change of the analyzer's state which needs to be persisted. Git repositories typically contain more than one branch. The \Lookout{} server notifies about pushes to each branch, and it is the responsibility of analyzers to correctly handle them. The typical behavior would be ignoring all the branches but the one which is marked as main on \Github{}. In most of the cases, it is the \texttt{master} branch, so we exercise \texttt{main} and \texttt{master} interchangeably below.

The high-level SDK takes care of technical issues, such as:

\begin{description}
    \item[Work with gRPC:] load balancing, connection pooling and related threading constraints;
    \item[Maintenance of the database with analyzer states:] the obvious way to achieve persistence. The states are stored in Advanced Scientific Data Format~\cite{asdf} with LZ4 compression on disk. The metadata about the states lives in any SQL database that is supported;
    \item[Caching states:] some repositories are going to send events much more frequently than others, and each database operation has a cost;
    \item[Logging:] logs include the request context with repository URL, revision, type of event, etc;
    \item[Metrics collection:] requests per time unit per analyzer type, elapsed time, various repository statistics, and failures. Metrics collection is especially important for machine learning-driven analyzers, because it is critical to measure various quality metrics at runtime to understand and improve the underlying models;
    \item[Common file filters:] removing certain files from consideration. Examples of such filters are binaries; very long lines (e.g. minified JavaScript); blacklisted file name prefixes (\texttt{node\_modules} in JavaScript, \texttt{vendor} in Go);
    \item[Common operations with file contents and UASTs:] \emph{e.g.} determining changed line numbers for a pair of files.
\end{description}

The high-level SDK defines the API for designing analyzers in Python.
It is enough to implement two functions ---~\texttt{train} and \texttt{analyze} which are invoked respectively on push and pull request events ---~to harness a fully featured analyzer.
For example, a basic analyzer that corrects typos in identifier names requires fewer than 100 lines of code\footnote{\url{https://github.com/src-d/lookout-sdk-ml/blob/master/lookout/core/examples/typos.py}}.

The high level SDK was chosen to create \StyleAnalyzer{} ---~the analyzer which trains on the master branch of a repository and suggests code formatting fixes.
\StyleAnalyzer's source code is available on \Github{}\footnote{\url{https://github.com/src-d/style-analyzer}} together with the related documentation and the datasets built for evaluation.
\section{Evaluation}
\label{sec:evaluation}

We evaluate \StyleAnalyzer{} on two benchmarks.
The first aims to measure how well \StyleAnalyzer{} models the style of a repository by applying a trained model on a held-out test dataset.
The predictions of the model are considered correct if they match the actual formatting elements in the original source code.
We carry out this evaluation on a collection composed of \numberTopRepos{} top-starred repositories from \Github{}.
We refer to that benchmark as the style modeling benchmark and report on the related findings in subsection~\ref{subsec:evaluation:style-modeling-benchmark}.
The second benchmark approximates the performance of \StyleAnalyzer{} in the wild by measuring how well it fixes style mistakes that we seeded manually in two large repositories.
We refer to it as the style defects fixing benchmark and devote to it subsection~\ref{subsec:evaluation:style-defects-fixing-benchmark}.

We trained all the models on a 32-core machine with 128 GB of memory running on Linux.
In practice, each model requires less than 5 GB of memory to be trained and less than 1 GB to be applied.
The pull request analysis always finishes in less than 15 min.

\subsection{Style modeling benchmark}
\label{subsec:evaluation:style-modeling-benchmark}

We select \numberTopRepos{} top starred open source \Github{} repositories which have JavaScript as their main language and ensure the diversity of their sizes to study how \StyleAnalyzer{} performs on both big and small codebases.
The head revisions of the selected repositories correspond to January 2019.
Among the largest repositories are \href{https://github.com/nodejs/node}{\texttt{nodejs/node}} and \href{https://github.com/facebook/react-native}{\texttt{facebook/react-native}}, which have more than one million lines; see Table~\ref{tab:valid-metrics} for further statistics.

\begin{figure}[h]
    \setcounter{figure}{\value{figurecounter}}
    \refstepcounter{figurecounter}
    \centering
    \includegraphics[width=\columnwidth]{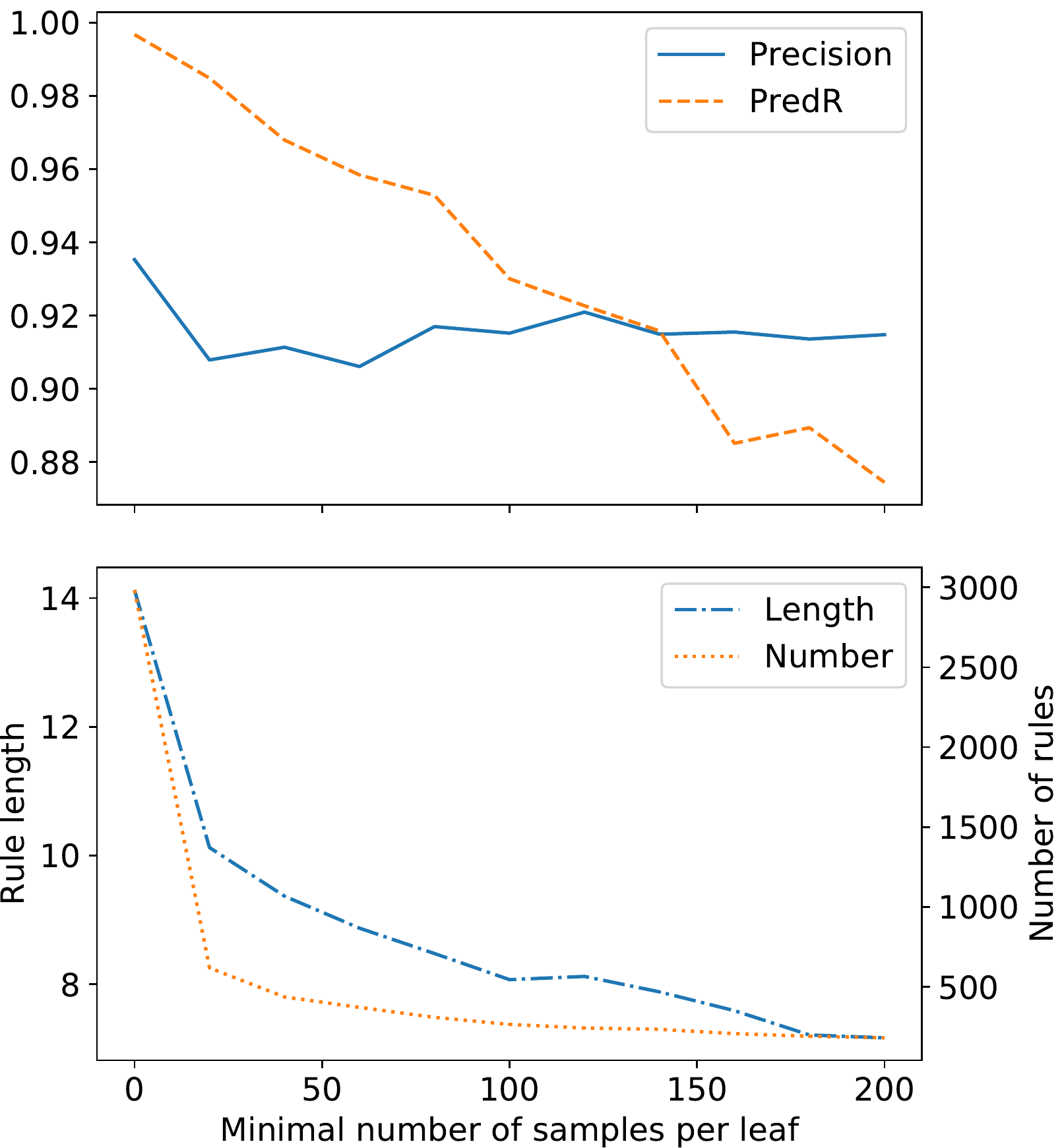}
    \vspace*{-1em}
    \caption{Impact of minimum samples per leaf on performance.}
    \vspace*{-0.8em}
    \label{fig:minimum-samples-leaf}
\end{figure}

\begin{table*}[t]
\begin{center}
    \begin{tabular}{rrrrrrrrrrr}
    \toprule
    repository     & precision   & PredR  &       recall & f1      & train samples        & LoC           & unique labels & rules & avg. rule length & training time, min \\
    \midrule 

node               &       0.965 &  0.951 &        0.918 &   0.941 &        \num{374298}  & \num{1074016} & 26 &        641     &           13.8 & 160 \\
webpack            &       0.957 &  0.956 &        0.915 &   0.936 &        \num{358012}  & \num{77731}   & 18 &        666     &           11.6 & 154 \\
meteor             &       0.900 &  0.845 &        0.761 &   0.825 &        \num{337627}  & \num{235411}  & 33 &        557     &           12.1 & 287 \\
react              &       0.943 &  0.974 &        0.919 &   0.931 &        \num{304465}  & \num{170920}  & 18 &        780     &           11.7 & 115 \\
atom               &       0.955 &  0.995 &        0.950 &   0.952 &        \num{265521}  & \num{137599}  & 16 &        440     &           11.1 & 125 \\
react-native       &       0.940 &  0.962 &        0.904 &   0.922 &        \num{264961}  & \num{131192}  & 22 &        693     &           13.7 & 206 \\
jquery             &       0.972 &  0.959 &        0.933 &   0.952 &        \num{197072}  & \num{55384}   & 19 &        391     &            9.5 & 82  \\
storybook          &       0.940 &  0.953 &        0.896 &   0.917 &        \num{161366}  & \num{43757}   & 15 &        494     &           10.3 & 39  \\
freeCodeCamp       &       0.928 &  0.960 &        0.891 &   0.909 &        \num{114020}  & \num{29044}   & 14 &        474     &            9.8 & 35  \\
express            &       0.937 &  0.979 &        0.918 &   0.928 &         \num{78411}  & \num{17460}   & 10 &        269     &            9.6 & 15  \\
30-seconds-of-code &       0.951 &  0.977 &        0.930 &   0.940 &         \num{67737}  & \num{11813}   & 10 &        151     &            7.1 & 8   \\
evergreen          &       0.894 &  0.958 &        0.857 &   0.875 &         \num{38387}  & \num{24507}   & 19 &         66     &           11.2 & 25  \\
citgm              &       0.936 &  0.933 &        0.873 &   0.904 &         \num{21941}  & \num{5349}    & 12 &         14     &            6.0 & 4   \\
axios              &       0.940 &  0.951 &        0.895 &   0.917 &         \num{21130}  & \num{7342}    & 10 &        143     &            7.3 & 4   \\
create-react-app   &       0.895 &  0.862 &        0.772 &   0.829 &         \num{16718}  & \num{14489}   & 12 &        101     &            6.8 & 4   \\
redux              &       0.937 &  0.844 &        0.791 &   0.858 &         \num{14783}  & \num{8963}    & 13 &         25     &            6.5 & 5   \\
reveal.js          &       0.897 &  0.842 &        0.755 &   0.820 &         \num{9974}   & \num{12926}   & 14 &         32     &            8.6 & 2   \\
carlo              &       0.878 &  0.931 &        0.817 &   0.846 &          \num{5529}  & \num{3449}    & 8  &         78     &            6.2 & 2   \\
telescope          &       0.806 &  0.570 &        0.460 &   0.585 &           \num{731}  & \num{467}     & 5  &          2     &            2.0 & 1   \\

\midrule
average            &       0.925 &  0.916 &        0.850 &   0.884 &        \num{139615}  &    ---~           & 15 &      317        &            9.2 & ---~ \\
weighted average   &       0.943 &  0.947 &        0.894 &   0.918 & ---~                 &    ---~           &  ---~  &---~             & ---~           & ---~ \\

\bottomrule
    \end{tabular}
    \end{center}
    \caption{Metrics measured on the validation part of the dataset. The last row is weighted by the number of samples.}
    \label{tab:valid-metrics}
    \vspace*{-1em}
\end{table*}

We make an assumption that the code style is consistent on file level, and we randomly split each repository's files into two groups. The size in bytes of the first group is 80\% of the overall size, of the second is 20\%. The files from the first group are used to train the model while the files from the second group are used for the validation benchmark.

To evaluate \StyleAnalyzer{}, we focus on both precision and prediction rate (PredR).
Precision indicates how noisy the predictions are. We believe that a score significantly lower than 95\% would jeopardize the users' trust in the model.
PredR, in turn, points out how often the model makes predictions.
Since we evaluate \StyleAnalyzer{} on real projects, it should be mentioned that inconsistent formatting across repositories could lead to poor precision.
Besides, the model does not always output a prediction for each labeled token because we remember that some rules can be disabled, see subsections \ref{subsec:methodology:decision-trees} and \ref{subsec:methodology:applying-rules}.
That is why we measure the percentage of predictions made:

$$
\text{PredR} = \frac{\mbox{number of predictions}}{\mbox{number of samples}}
$$

In practice, the label frequency distribution has a long tail. Thus we measure the frequencies of compound labels and exclude the rare ones by putting a threshold of 80 occurrences throughout all the files. According to our measurements, this value retains approximately 30\% of the labels and reaches a decent compromise between keeping significant labels and cutting away outliers.
Regarding the confidence threshold whose goal is to select the most relevant rules, see subsection \ref{subsec:methodology:decision-trees}, we experimentally found 0.92 to yield good results; it keeps from 10\% to 50\% of the rules depending on the dataset and model.
To further reduce the number of rules ---~and to regularize the model~--- we set a minimum number of samples for each leaf of the tree to a threshold that we determine experimentally.
Figure~\ref{fig:minimum-samples-leaf} shows the evolution of the precision and PredR on one side, and the number of rules and their average length on the other side, both over the minimum number of samples per leaf, on the training set.
The number of trees in the random forest was fixed to 10.
After analyzing the results, a good trade-off between high PredR and the interpretability of the rules is to set the optimal value of the minimum number of samples per leaf to \minSamplesLeaf; we will keep this value in the next experiments.

Table~\ref{tab:valid-metrics} gathers the model's metrics for the \numberTopRepos{} top starred JavaScript repositories on which we achieve 94\% precision at 95\% PredR on weighted average. \StyleAnalyzer{} performs poorly on small codebases, as can be observed in Figure~\ref{fig:precision-size}. We relate this fact to not having a sufficient number of training samples. Furthermore, the precision on \href{https://github.com/segmentio/evergreen}{\texttt{evergreen}} is low because it contains a mixture of JavaScript and JSX with different formatting.

We also note that repositories with strong style guidelines ---~such as \href{https://github.com/jquery/jquery}{\texttt{jQuery}}~--- are easier to model and produce simpler models (with less rules) by a significant margin compared to repositories with inconsistent styles ---~such as \href{https://github.com/freeCodeCamp/freeCodeCamp}{\texttt{FreeCodeCamp}}.

\begin{figure}[b]
    \setcounter{figure}{\value{figurecounter}}
    \refstepcounter{figurecounter}
    \centering
    \vspace*{-0.5em}
    \includegraphics[width=\columnwidth]{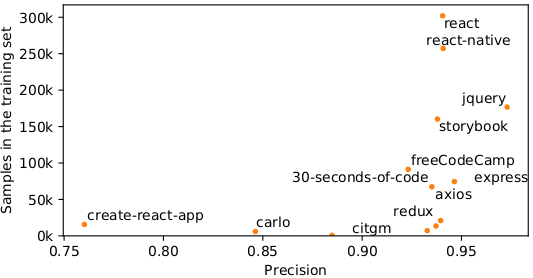}
    \caption{Effect of the number of samples in the training set on precision.}
    \vspace*{-0.5em}
    \label{fig:precision-size}
\end{figure}

We have additionally measured the ratio of rules which contain parental attribute comparisons, i.e. leverage the structural information: 60\%-90\% depending on the repository, which proves its usefulness. UAST difference check cuts 8\% predictions on average across 8,000 files, which grants a considerable precision gain. Attributes pruning removes up to 60\% comparisons in our 19 repositories, which leads up to 55\% less rules without any precision drop.

The style modeling benchmark gives a general idea of how good the model can approximate the style of a repository. It does not directly demonstrate the quality of suggestions for fixing style defects in source code.

\subsection{Style defects fixing benchmark}
\label{subsec:evaluation:style-defects-fixing-benchmark}

To simulate \StyleAnalyzer{}'s real usage, we manually compiled a dataset of artificial style mistakes in the form of commits that deliberately introduce formatting inconsistencies in JavaScript files of some \Github{} repositories.
To exploit this dataset, we first train the model on the original revision and then apply those commits on top and study how the mined rules perform. We add a single format distortion per file to simplify the fix correctness check (see Figure~\ref{fig:artificial-style-mistakes}). There are 80 whitespace and 60 newline insertions and removals, 10 indentation changes and 20 different pairs of quotes; 170 changes overall.

\begin{figure}[t]
    \setcounter{figure}{\value{figurecounter}}
    \refstepcounter{figurecounter}
    \begin{subfigure}[b]{0.5\textwidth}
        \begin{lstlisting}[frame=tb, framexrightmargin=-0.03\textwidth, mathescape=true, language=JavaScript]
    function classesToArray( value ) {
    	if ( isArray( value ) ) {
    	    return value;
    	}
    	if ( typeof value === "string" ) {
    		return value.match( rnothtml ) || [];
    	}
    	return [];
    }
        \end{lstlisting}
        \vspace*{-0.9em}
        \captionof{figure}{}
        \vspace*{0.4em}
    \end{subfigure}
    \begin{subfigure}[b]{0.5\textwidth}
        \begin{lstlisting}[frame=tb, framexrightmargin=-.03\textwidth, mathescape=true, language=JavaScript]
    function classesToArray(value) {
    	if ( isArray(value) ) {return value;}
    	if ( typeof value === 'string' ) {
    	    return value.match(rnothtml) || [];
    	}
    	return [];
    }
        \end{lstlisting}
        \vspace*{-0.9em}
        \captionof{figure}{}
        \vspace*{0.4em}
    \end{subfigure}
    \caption[figure]{(a) Example of JavaScript code and (b) its modified version that includes style inconsistencies: (i) spaces around API calls have been removed (ii) a tabulation has been replaced by a four spaces (iii) the first \texttt{if} statement has been shortened into one line (iv) single quotes have replaced double quotes}
    \vspace*{-1em}
    \label{fig:artificial-style-mistakes}
\end{figure}

We sort the rules by confidence in decreasing order and watch quality metrics as we successively enable less and less confident rules.
Incrementing the number of rules understandably increases the overall number of predictions. Yet the number of mistakes grows, too.
Therefore, we are interested in the precision and PredR metrics: (i) precision equals the ratio of correctly predicted fixes over the total number of predictions made, and (ii) PredR equals the ratio of the total number of predictions made over the ground truth number of added mistakes. The results are shown on Figure~\ref{fig:noisy-plots}.

\begin{figure}[t]
    \setcounter{figure}{\value{figurecounter}}
    \refstepcounter{figurecounter}
    \centering
    \vspace*{-0.2em}
    \includegraphics[width=\columnwidth]{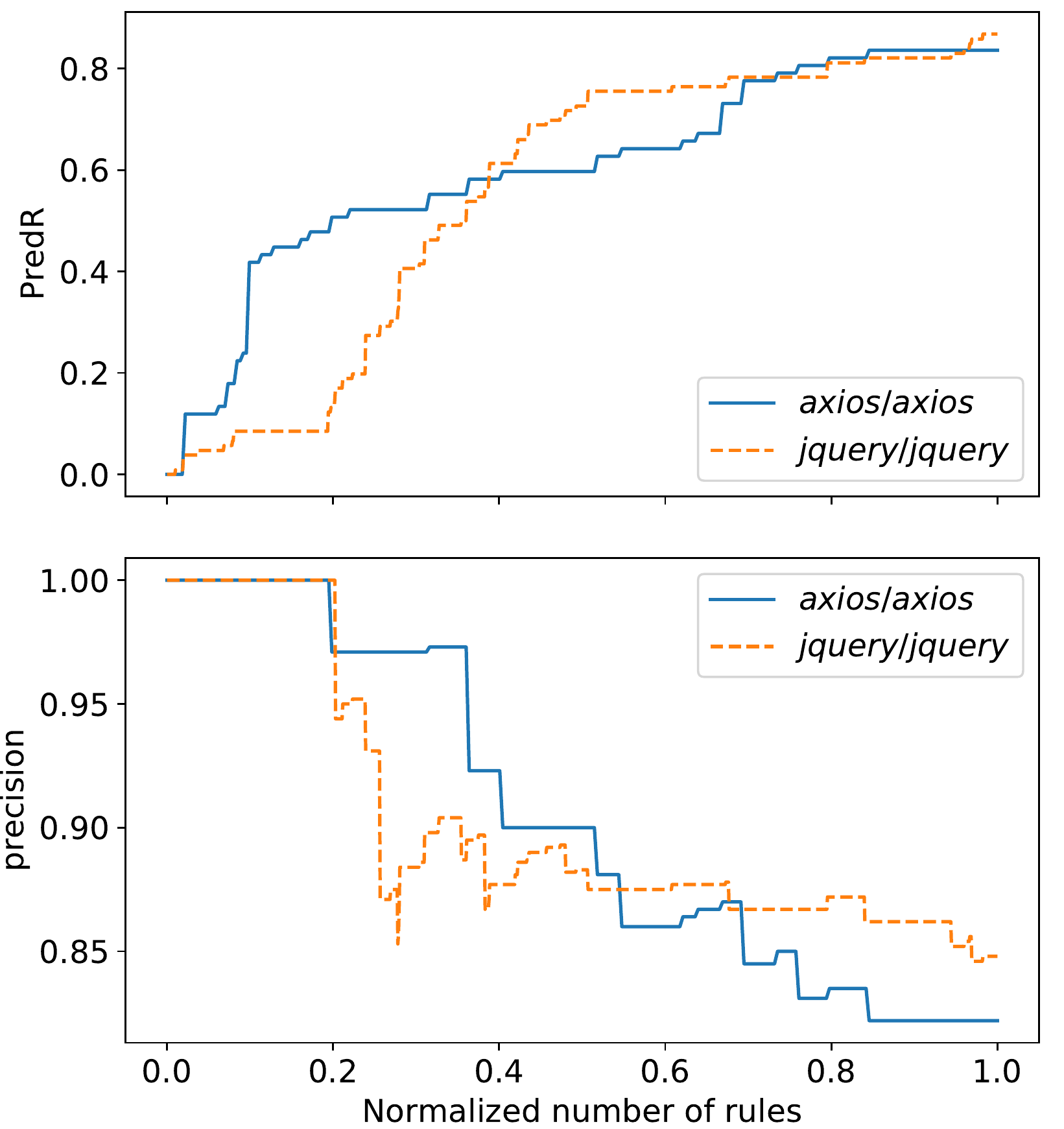}
    \vspace*{-1em}
    \caption{Metrics variance on the dataset of style defects: (a) PredR and (b) precision evolutions over the number of rules retained by the model. The rules are sorted by confidence and their overall amounts are respectively \numberRulesAxios and \numberRulesJquery elements for the \textit{axios/axios} and \textit{jquery/jquery} \Github{} repositories.}
    \vspace*{-0.8em}
    \label{fig:noisy-plots}
\end{figure}

We use the style defects fixing benchmark to determine the lowest rule confidence threshold to stay above the target 95\% precision. 
The plots show that 95\% precision corresponds to roughly 40\% of all the rules and 60\% PredR. That is equivalent to 92\% rule confidence threshold.
The latter number is less than 95\% because more confident rules tend to trigger more frequently.

\subsection{Rules visualization}
\label{subsec:evaluation:rules-visualization}

\begin{figure*}[t]
\setcounter{figure}{\value{figurecounter}}
\refstepcounter{figurecounter}
\centering
\includegraphics[width=1\linewidth]{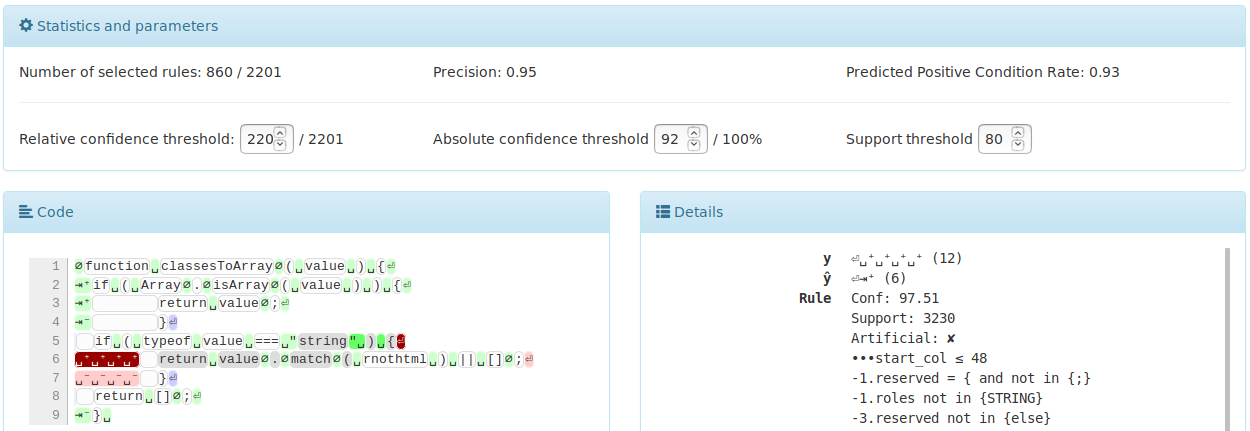}
\captionof{figure}{A screenshot of the web based application to visualize any annotated source code and the triggered rules.}
\vspace*{-1em}
\label{fig:visualizer}
\end{figure*}

We developed a web based application to visualize how a trained model works.
This application enables users to load any JavaScript file, apply the mined rules and inspect each triggered position in the code (see Figure~\ref{fig:visualizer}).
We support changing some of the rules' optimization parameters, such as the confidence and support thresholds.
The triggered rule is displayed on the right of the visualizer's window as a logical conjunction of attribute clauses.
We also print its confidence and support, \emph{i.e.} how many times the rule has been fired during training.
Color coding is applied to make it easier to read a snippet of code with the corresponding model's predictions. More precisely, we use the following colors:

\vspace*{0.5em}
\begin{tabular}{M{0.1cm} l}
\green{ } & the model's prediction and the input label match \tabularnewline
\red{ } & the model's prediction and the input label differ  \tabularnewline
\blue{ } & the model is disabled on this sample
\end{tabular}
\vspace*{0.5em}

This web application turns out to be particularly useful when debugging the models ---~when we need to understand their decisions or to unveil problems during feature extraction.
However, it is not able to show the actual generated source code for now.
\section{Future work}
\label{sec:future-work}

As shown by the evaluation performed in section~\ref{sec:evaluation}, the formatting style can differ across a single repository. The way the current model works is to average all the examples and infer a unified solution. A promising alternative is to cluster files which are written in a similar style together and train several models on those clusters. Choosing the right model for new files would then be a challenge. A possible solution would be to apply all of them and take the least contradictory.

It would also be beneficial to present the mined rules with source code. Then the model becomes directly editable. Similar attribute comparisons should be grouped together, however, finding the optimal grouping imposes a separate problem. Besides, we would have to recompute the confidence of the edited rules.

\StyleAnalyzer{} does not require any configuration, so it can be run on many repositories at scale. It would be interesting to study the common mined code formatting patterns across \Github{}, although that would require improving the rule matching described in subsection~\ref{subsec:methodology:user-feedback}. Moreover, it would be possible to transfer style, for example, train formatting rules on a set of trusted codebases which follow best practices and apply them to other projects. Apart from the global style transfer, partial formatting should be explored as well.

It is promising to investigate Meta-learning approaches to training in the future~\cite{vilalta2002perspective}.
Meta-learning can be summarized as principled methods that exploit meta-knowledge to obtain efficient models and solutions by adapting machine learning and data mining processes.
The training would be divided into two distinct stages.
The first stage and also the most time-consuming one would be to train style embeddings by asking a model to reconstruct the formatting of a codebase with only ASTs and a style embedding at its disposal.
The second stage would be to adapt this knowledge to model the style of a particular repository: this would entail finding the optimum embedding to construct the formatting of the existing codebase given its ASTs.
Since the embeddings are the entities in which the model condenses all the style information, it becomes possible to exploit well-known techniques from both Natural Language Processing and Computer Vision areas to find the best embeddings to model the desired style~\cite{rusu2018meta}.
The speculated two-stage training process is sound because the model architecture can be arbitrarily complex and expressive.
It can potentially learn how certain styles are applied in certain corner cases in millions of repositories, and produce predictions for a single repository, since the core of what varies is shared.

\section{Conclusion}
\label{sec:conclusion}
We presented \StyleAnalyzer, a novel open source tool that automatically fixes code formatting defects during code reviews.
It learns human-readable rules of formatting from a codebase using the decision tree forest model and does not require any human participation during its training. Application of the mined rules to code in pull requests generates code suggestions which improve stylistic consistency. The achieved model performance reached the level of practical usability.
Our experiments with popular JavaScript repositories on \Github{} showed that \StyleAnalyzer{} stays effective at diverse training set sizes, reaching both a significant percentage of predictions made (PredR) and a high precision score.
Besides, we pointed out the positive performance of our tool on a manually created dataset of formatting style mistakes. We additionally demonstrated the web application that we developed to visualize how rules fire at different places in analyzed code.

\StyleAnalyzer{} is based on \Lookout, a featureful framework for assisted code review. \Lookout{} allowed us to focus on more important tasks and develop code analyses faster. We hope that our framework will inspire other researchers to conduct assisted code review studies in the future.
We release \StyleAnalyzer{} as a reusable and extensible software package together with the datasets used for evaluation.
We plan to turn \StyleAnalyzer{} into a commercial product and improve the suggestions quality by analyzing how real users interact with the tool.
\section{Acknowledgment}
\label{sec:acknowledgment}
We warmly thank source\{d\} Applications Team members, namely Maxim~Sukharev, Carlos~Martin, Alexander~Bezzubov, David~Pordomingo and Lou~Marvin~Caraig, for their hard work and diligent support on the \Lookout{} framework.

\bibliographystyle{IEEEtran}
\bibliography{generic}

\end{document}